\documentclass[10pt,twocolumn,letterpaper]{article}

\usepackage{cvpr}              % To produce the CAMERA-READY version
\usepackage{graphicx}
\usepackage{amsmath}
\usepackage{amssymb}
\usepackage{booktabs}
\usepackage{multirow}
\usepackage{tabularx}
\usepackage{lipsum}

\usepackage{array}

\newcolumntype{Y}{>{\centering\arraybackslash}m{0.95cm}} % adjust width as needed

\newcolumntype{Z}{>{\centering\arraybackslash}m{1.5cm}} % adjust width as needed

\usepackage[pagebackref,breaklinks,colorlinks]{hyperref}

\usepackage[capitalize]{cleveref}
\crefname{section}{Sec.}{Secs.}
\Crefname{section}{Section}{Sections}
\Crefname{table}{Table}{Tables}
\crefname{table}{Tab.}{Tabs.}

\def\cvprPaperID{*****} % *** Enter the CVPR Paper ID here
\def\confName{CVPR}
\def\confYear{2023}

\begin{document}

\title{Benchmarking the Robustness of UAV Tracking Against Common Corruptions}

\author{Xiaoqiong Liu$^{1}$ \;\;\; Yunhe Feng$^{1}$ \;\;\;  Shu Hu$^{2}$  \;\;\; Xiaohui Yuan$^{1}$ \;\;\; Heng Fan$^{1}$\\
$^{1}$Department of Computer Science and Engineering, University of North Texas\\
$^{2}$Department of Computer and Information Technology, Purdue University\\
{\tt\small xiaoqiongliu@my.unt.edu; heng.fan@unt.edu}
% \and
}
\maketitle

%%%%%%%%% ABSTRACT
\begin{abstract}

The robustness of unmanned aerial vehicle (UAV) tracking is crucial in many tasks like surveillance and robotics. Despite its importance, little attention is paid to the performance of UAV trackers under common corruptions due to lack of a dedicated platform. Addressing this, we propose \textbf{\emph{UAV-C}}, a large-scale benchmark for assessing robustness of UAV trackers under common corruptions. Specifically, UAV-C is built upon two popular UAV datasets by introducing 18 common corruptions from 4 representative categories including adversarial, sensor, blur, and composite corruptions in different levels. Finally, UAV-C contains more than 10K sequences. 
To understand the robustness of existing UAV trackers against corruptions, we extensively evaluate 12 representative algorithms on UAV-C. Our study reveals several key findings: 1) Current trackers are vulnerable to corruptions, indicating more attention needed in enhancing the robustness of UAV trackers; 2) When accompanying together, composite corruptions result in more severe degradation to trackers; and 3) While each tracker has its unique performance profile, some trackers may be more sensitive to specific corruptions. By releasing UAV-C, we hope it, along with comprehensive analysis, serves as a valuable resource for advancing the robustness of UAV tracking against corruption. Our UAV-C will be available at \url{https://github.com/Xiaoqiong-Liu/UAV-C}.

\end{abstract}

%%%%%%%%% BODY TEXT
\section{Introduction}
\label{sec:intro}

\begin{figure}
  \includegraphics[width=\linewidth]{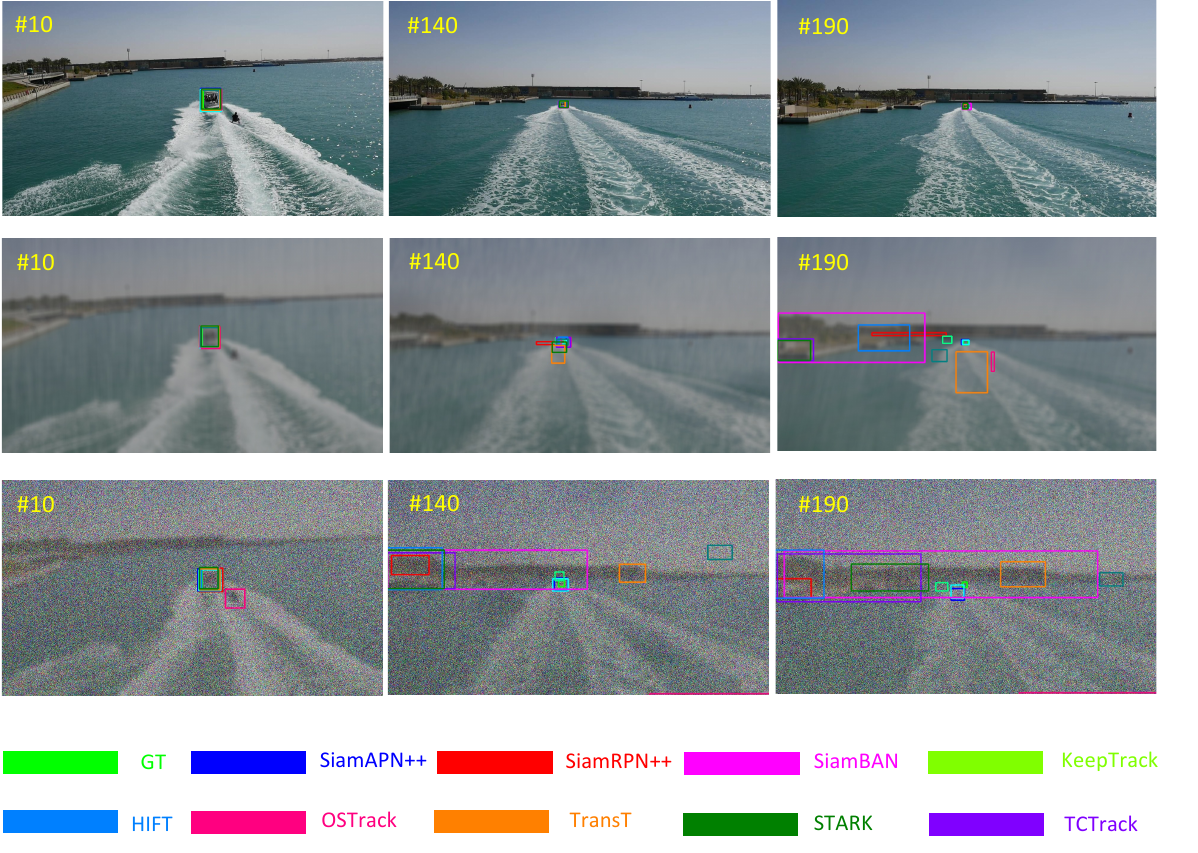}
  \caption{Degradation of UAV trackers under several corruptions (only partial of corruption is shown due to space limitation). The first row displays clean frames for reference. The second row illustrates the impact of Rain-D corruption, while the third row showcases the effects of Rain-D-G corruption. Best viewed in color.}
  \label{visres}
  \label{fig:teaser}
\end{figure}

Unmanned aerial vehicle (UAV) tracking aims to localize the target given its initial status (\eg, a bounding box) in a sequence captured by a drone. It has numerous crucial applications including video surveillance, robotics, and smart agriculture. Despite considerable progress~\cite{li2020autotrack, cao2021hift, cao2021siamapn++, li2023adaptive, cao2022tctrack, Yao2023SGDViTSD, Li_2023_ICCV}, current UAV tracking mainly focuses on improving performance under normal scenarios, yet neglects the robustness of trackers under abnormal scenes with various corruptions such as weather and sensor that may occur in natural drone-captured videos, mainly due to the lack of a dedicated data platform.

To address this, we make the first attempt by introducing a large-scale and diverse benchmark dedicated to the study of UAV tracking robustness under common corruptions. In specific, our proposed benchmark, dubbed \emph{\textbf{UAV-Corruption}} or \emph{\textbf{UAV-C}}, is built upon two popular benchmarks UAV-123 (10fps) \cite{mueller2016benchmark} and DTB-70~\cite{Li2017VisualOT}, and introduces 18 types of synthesized corruptions. These corruptions cover both distinct and complicated scenarios encountered during UAV operations and can be grouped into 4 categories, including weather, blur, sensor, and composite corruptions, and each of them contains three different levels. Notably, several of these corruptions are tailored for UAV video sequences, including depth-aware fog and continuous zoom blur. In total, UAV-C consists of 9843 videos with more than 3.1 million frames. By developing UAV-C, we aims to provide a dedicated evaluation platform that can be used for assessing UAV tracking robustness against different corruptions.

In order to understand the performance of existing UAV trackers on UAV-C, we extensively evaluate 12 representative tracking algorithms, which are diverse in terms of pre-training strategies, feature extractors, and relation modeling. The evaluation results reveal a few key findings. \emph{First}, the tracking performance is generally degraded in presence of corruptions (see Fig.~\ref{fig:teaser}). For example, the performance of the evaluated has a significant drop, with precision dropping between 2.5\% and 72.9\%., as analyzed later in experiments. \emph{Second}, when different corruptions happen simultaneously, \ie, composite corruptions, the trackers are further degenerated. \emph{Third}, different tracking algorithms may be sensitive to different corruptions, which shows specific designs may be needed to deal with corruptions for certain trackers. With our extensive assessment, we show that, corruption is a severe challenge to robust UAV tracking, and more efforts are needed to enhance the robustness in the future.

In summary, our contributions are: (i) We propose UAV-C, a large-scale benchmark for evaluating UAV tracking robustness again corruptions; (ii) We assess 12 representative UAV trackers to understand existing methods on UAV-C and to offer baseline for comparison; and (iii) We conduct in-depth analysis on UAV-C that provides guidance for future development of more robust tracking algorithms.

\section{Related Work}

In this section, we mainly focus on  benchmarks that are relevant to our work. For UAV related tracking algorithms, please kindly refer to a recent survey~\cite{fu2023siamese}.

\subsection{UAV Tracking Benchmark}

In recent years, UAV tracking has witnessed significant advancements, driven by the development of comprehensive benchmarks. \textbf{UAV123}~\cite{mueller2016benchmark} proposes the first benchmark exclusively crafted for UAV tracking tasks, consisting of over 100 video sequences. It incorporates data from both professional-grade and consumer-grade UAVs, along with simulator-generated data. \textbf{DBT70}~\cite{Li2017VisualOT} offers a collection of 70 video sequences procured from diverse sources, ranging from drones to YouTube recordings. This dataset is enriched with manually annotated bounding boxes, specifically targeting pedestrians and vehicles. \textbf{VisDrone}~\cite{zhu2021detection}, a recent large-scale dataset, provides a substantial volume of images paired with meticulous annotations. Continuously expanded and updated, VisDrone encompasses diverse environmental conditions, including various weather scenarios (\eg, sunny, cloudy, and rainy), different altitudes, and varying camera viewpoints. \textbf{Anti-UAV}~\cite{jiang2021anti} features videos depicting different types of UAVs navigating through diverse lighting conditions—both day and night—and employing various light modes (infrared and visible) against different backgrounds. This dataset is meticulously curated to ensure diversity for effective tracking purposes. \textbf{UAVDT}~\cite{Du2018TheUA} compiles 100 video sequences captured from a UAV platform navigating urban areas, including scenes like highways and T-junctions. Annotated for tracking tasks, UAVDT provides object-bounding boxes to facilitate comprehensive analysis. \textbf{MDOT}~\cite{zhu2020multi} is recently proposed for multi-drone single-object tracking, presenting video clips captured concurrently by two or three drones tracking the same target during different daytime scenarios. \textbf{BioDrone}~\cite{zhao2023biodrone} is recently introduced to include videos captured by a flapping-wing bionic drone, inducing pronounced camera shake. This distinctive dataset emphasizes tracking minuscule targets amidst substantial frame changes, serving as a novel benchmark for robust tracking.

Different from the above UAV tracking benchmarks that often focuses on tracking under normal scenarios, UAV-C has a different goal by aiming at localizing targets under the corrupted scenes, which expects to facilitate UAV tracking from a difference perspective and bring more attention to the research on robustness against corruptions.

\subsection{Other Corruption-Related Benchmarks}
 Recent studies show deep learning models have poor generalization~\cite{Geirhos2018GeneralisationIH} and exhibit vulnerabilities to adversarial examples~\cite{Goodfellow2014ExplainingAH,Szegedy2013IntriguingPO,liu2023transferable}, common corruptions~\cite{hendrycks2019benchmarking}, etc. A promising avenue involves the generation of real-world corruptions on clean datasets to serve as a benchmark for assessing model robustness. Drawing inspiration from methodologies like ImageNet-C \cite{hendrycks2019benchmarking}, initially introduced for image classification, this approach introduces 15 corruption types encompassing factors such as noise, blur, weather, and digital corruption. This methodology has been extended to various tasks and scenarios~\cite{Michaelis2019BenchmarkingRI,Kamann2019BenchmarkingTR}. 

These benchmarks play a crucial role in understanding how well models can perform under less-than-ideal circumstances. Our UAV-C shares a similar spirit with these datasets, but is different in that we specially focus on the task of UAV tracking under corruptions. In addition, we design the novel composite corruptions yet others often contain only single corruptions.

\section{ Proposed UAV-C Benchmark }

To enhance robustness and applicability of UAV tracking in diverse scenarios, we introduce the UAV-C benchmark, as described in the following, which simulates a range of realistic corruptions that UAVs may encounter. % This section is structured as follows: We first describe the data sources utilized for this benchmark, followed by a detailed account of the corruption types incorporated. Finally, we outline the construction process of the UAV-C benchmark.

\renewcommand{\arraystretch}{1.1}
\begin{table*}[!t]
  \centering 
  \resizebox{\linewidth}{!}{
  \begin{tabular}{@{}lll@{}}
    \hline
    \textbf{Corruption Type} & \textbf{Description} & \textbf{Severity Levels} \\
    \hline
    \textbf{Weather} & Fog, Rain, Snow, Frost, and Spatter & Low, Medium, High \\
    \hline
    \textbf{Sensor} & Gaussian, Poisson, Impulse, Speckle noises and Low Contrast & Low, Medium, High \\
    \hline
    \textbf{Blur} & Defocus, Motion, Zoom, and Gaussian Blur & Low, Medium, High \\
    \hline
    \textbf{Composite} & Dual Interaction Fusion (DIF), Tri-Interaction Synthesis (TIS) & Low, Medium, High \\
    \hline
  \end{tabular}}
  \caption{Summary of 18 corruptions from 4 representative categories in UAV-C. Similar to~\cite{hendrycks2019benchmarking}, each corruption has different levels.}
  \label{tab:corrup}
\end{table*}

\begin{figure*}
    \centering
    \includegraphics[width=0.95\linewidth]{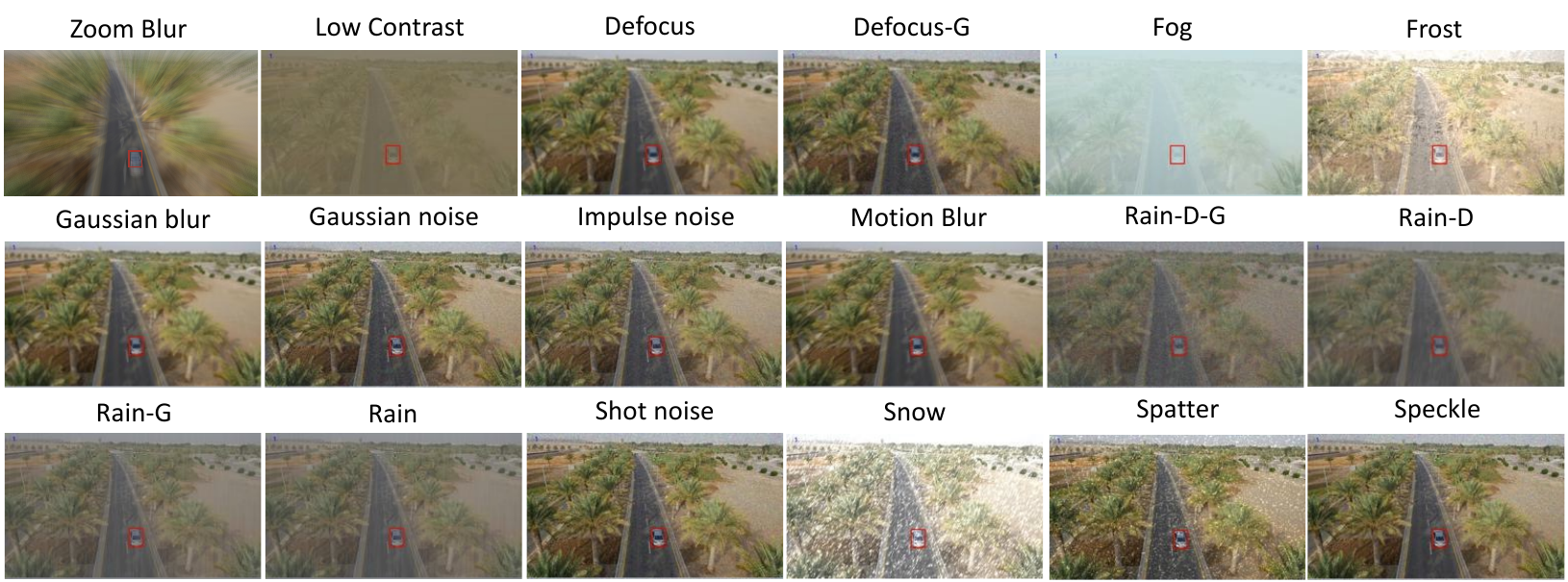}
    \caption{Illustration of the clean frame and its 18 corrupted generated using our method. }
\label{fig2}
\end{figure*}

\subsection{Data Source}
To build UAV-C, we use two prominent UAV datasets: UAV123-10fps~\cite{mueller2016benchmark} and DBT70~\cite{Li2017VisualOT}. UAV123-10fps comprises 123 sequences captured from diverse scenes recorded at 10 frames per second, resulting in a comprehensive coverage of common UAV tracking challenges. DBT70, on the other hand, includes 70 high-resolution videos that provide a wide array of dynamic backgrounds and fast-moving objects. Eventually, the combined dataset provides 193 videos  across various landscapes and conditions, serving as a robust foundation for evaluating UAV tracking robustness.

\subsection{Corruption Categories}

UAV-C introduces four primary types of corruptions, including weather, sensor, blur, and composite corruptions, with each mimicking real challenges for UAV tracking.

\textbf{Weather Corruption} simulates adverse weather conditions including \emph{fog}, \emph{rain}, \emph{snow}, \emph{frost}, and \emph{spatter}, affecting visibility and tracking accuracy.

\textbf{Sensor Corruption} represents common sensor-related noises, including \emph{Gaussian}, \emph{Poisson}, \emph{Impulse}, \emph{Speckle} noises, and \emph{Low Contrast}, that can occur due to hardware limitations or environmental factors.

\textbf{Blur Corruption} includes motion and focus blur to account for rapid UAV movements and autofocus discrepancies, which are prevalent in dynamic tracking scenarios.

\textbf{Composite Corruption} is specially designed in our benchmark, by combining compositions from the weather, sensor, and blur categories to explore the complex interactions and compounded effects on tracking performance.

Each corruption type includes several specific instances, detailed in Tab.~\ref{tab:corrup}, with definitions and three severity levels ranging from low to high to cover a broad spectrum of potential disruptions, similar to~\cite{hendrycks2019benchmarking}.

\subsection{Construction of UAV-C}

Building upon the defined corruptions and sourced data, UAV-C is meticulously constructed by applying the 18 identified corruptions across 51 severity levels to video frames from UAV123-10fps and DBT70 datasets. This process results in a comprehensive and challenging dataset, dubbed UAV-C, which is designed to rigorously evaluate and enhance the resilience of UAV tracking algorithms. We visualize different corruptions under one certain level in Fig.~\ref{fig2}, providing insights into the expected challenges and facilitating a deeper understanding of the benchmark's complexity.

In the following, we delve into the specifics of each corruption type, elucidating the rationale behind their selection, the methodology employed for their simulation, and their anticipated effects on UAV tracking performance.

\textbf{Weather-level corruptions.} Weather conditions pose a significant threat to the robustness of algorithms relying on drone-collected data \cite{Mokayed2023NordicVD}. We study four distinct weather-related corruptions, including fog, rain, snow, and frost. Among them, fog and heavy rain corruption can diminish visibility of the targets, making them hard for UAV trackers and leading to degradation. In addition, the falling snow further exacerbates visibility issues, while frost accumulation on sensors or lenses can impair their functionality. All of them will cause difficulties for UAV tracking.

To simulate these adverse weather scenarios, we employ algorithm-based augmentation with three severity levels for each weather condition. Recognizing the depth variation inherent in drone-collected data, we introduce a depth-aware approach~\cite{Ranftl2019TowardsRM}, specifically tailored for simulating fog.

\textbf{Sensor-level corruptions.} Sensors, influenced by diverse internal or external factors such as lighting conditions~\cite{hendrycks2019benchmarking} and reflective materials, can introduce various forms of corruptions. We consider five major sensor-level corruptions, including Gaussian noise, Poisson noise, Impulse noise, Speckle noise, and Contrast. Gaussian Noise, Uniform Noise, and Impulse Noise simulate visual noise patterns arising from low-lighting conditions or camera defects~\cite{hendrycks2019benchmarking}. Contrast levels may vary, influenced by lighting conditions and the color of the photographed object~\cite{hendrycks2019benchmarking}.

\textbf{Blur corruptions.}
 The efficacy of UAV tracking is intricately tied to the excellence of images or frames captured by the onboard cameras. Several factors can contribute to image blurring. Defocus blur emerges when an image is not in sharp focus. During swift UAV movements, particularly in rapid changes of direction or high-speed flights, motion blur may manifest in the captured images. Zoom blur materializes when a camera rapidly approaches an object. Additionally, in certain instances, post-processing filters or algorithms applied for image quality enhancement or noise reduction may unintentionally introduce Gaussian blur if not meticulously tuned.

\textbf{Composite corruptions.} Besides single type of corruption, we also consider composite that combines multiple corruptions. Specifically, we design two types of composite corruptions, including dual interaction fusion (DIF) and tri-interaction synthesis (TIS). DIF involves the synergistic impact of two interactions between weather and sensor or sensor and blur. TIS represents the intricate synthesis of three interactions, incorporating the combined effects of weather, sensor, and blur corruptions. In UAV-C, we introduce 3 DIFs and 1 TIS. Specifically, the composite DIF corruption, denoted as $\mathcal{P}_{\text{A-B}}$, is defined by applying the B corruption to the A-corrupted video frame as $\mathcal{P}_{\text{{A-B}}} = \text{{B}} \circ \mathcal{P}_{\text{{A}}}$. Similarly, the composite TIS corruption, denoted as $\mathcal{P}_{\text{A-B}}$, is obtained by applying the C corruption to the A-B-corrupted video frame as $\mathcal{P}_{\text{{A-B-C}}} = \text{{C}} \circ \mathcal{P}_{\text{{A-B}}}$. Concretely, we consider Rain-Defocus (Rain-D), Rain-Gaussian noise (Rain-G), and Defocus-Gaussian noise (Defocu-G) for DIF corruptions, and Rain-Defcous-Gaussian noise for TIS corruption.

\renewcommand{\arraystretch}{1.15}
\begin{table*}[!t]
\centering
\resizebox{\textwidth}{!}{%
    \large % Set the font size to \large
    \begin{tabular}{@{}c|c|cccc|cccccc|cc@{}}
    \hline
    \multicolumn{2}{c|}{\multirow{1}{*}{\textbf{Corruption}}} & \multicolumn{4}{c|}{\textbf{CNN-based}} &  \multicolumn{6}{c|}{\textbf{CNN-Transfomer-based}}  & \multicolumn{2}{c}{\textbf{One Stream Transformer}}\\
     \multicolumn{2}{c|}{($mS_{c}\uparrow$)}& SiamAPN++ & SiamBAN & SiamRPN++ & KeepTrack & ToMP  & HIFT & TCTrack & TransT & STARK & SGD &Aba-ViTrack & OSTrack\\
    \hline\hline
    \multicolumn{2}{c|}{\textbf{None} ($S_{clean}$)} & 57.9 & 61 & 60.4  & 67.7 & 67.1 & 57.3 & 59.4 & 65.5 & 66.7 & 59 & 65.8 & 69.3 \\
    \hline
    \multirow{4}{*}{\textbf{Weather}} & \textbf{Snow} & 47.7 & 54.5 & 53.4 & 60.0 & 62.0  & 48.1 & 49.6 & 59.5 & 60.1 & 47.7 & 59.8 & 63.5 \\
    &\textbf{Frost} & 50.8 & 53 & 51.9 & 60.2 & 61.1  & 51.1 & 51.3 & 59.1 & 60.8 & 52.0 & 59.7 & 63.9 \\
    &\textbf{Fog} & 37.3 & 51.8 & 52.9 & 59.4 & 60.1  & 41.9 & 44.9 & 59.7 & 57.7 & 43.7 & 49.3 & 61.7 \\
    &\textbf{Rain} & 54.0 & 59.2 & 58.9 & 64.6 & 65.9 & 56.4 & 57.1 & 64.0 & 65.2 & 54.8 & 64.7 & 68.2 \\
    &\textbf{Spatter} & 52.4 & 57.1 & 56.1 & 61.6 & 62.5  & 51.7 & 52.7 & 60.7 & 62.2 & 53.0 & 61.2 & 66.2  \\
    \hline
    \multirow{5}{*}{\textbf{Sensor}} & \textbf{Gaussian Noise} & 41.1 & 50.9 & 49.5 & 54.7 & 56.6  & 46.1 & 39.2 & 54.2 & 55.2 & 43.3 & 55.3 & 58.7 \\
    &\textbf{Shot Noise} & 43.5 & 52.8 & 51.5 & 55.6 & 58.5 & 46.4 & 41.1 & 56.0 & 56.9 & 44.8 & 57.3 & 59.9 \\
    &\textbf{Impulse Noise} & 41.5 & 52.7 & 50.7 & 55.7 & 58.5  & 46.3 & 40.4 & 56.3 & 56.3 & 43.5 & 56 & 61.3 \\
    &\textbf{Speckle Noise} & 48.0 & 54.7 & 54.2 & 58.7 & 61.3 & 51.0 & 47.8 & 59.3 & 59.5 & 49.1 & 59.4 & 63.0 \\
    &\textbf{Low Contrast} & 37.0 & 51.4 & 54.5 & 60.5 & 60.9  & 41.4 & 45.5 & 61.4 & 59 & 40.9 & 52.4 & 62.0 \\
    \hline
    \multirow{4}{*}{\textbf{Blur}} & \textbf{Defocus} & 49.3 & 53.1 & 53.6 & 58.3 & 59.4  & 49.9 & 48.8 & 58.3 & 57.4 & 48.7 & 55.9 & 60.5 \\
    &\textbf{Motion} & 47.0 & 51.8 & 50.6 & 54.4 & 54.1  & 48.3 & 49.9 & 54.8 & 52.6 & 47.4 & 51.4 & 56.4 \\
    &\textbf{Zoom} & 18.3 & 21.0 & 22.9 & 21.7 & 22.1  & 20.8 & 21.1 & 14.4 & 20.5 & 19.6 & 18.1 & 20.4 \\
    &\textbf{Gaussian Blur} & 48.9 & 52.8 & 54 & 59.4 & 58.4 & 49.9 & 48.9 & 58.6 & 57.6 & 48.2 & 56.6 & 61.1 \\
    \hline
    \multirow{4}{*}{\textbf{Composite}} & \textbf{Rain-D} & 42.4 & 49.8 & 51.0 & 55.7  & 57.3 & 45.3 & 45.4 & 55.8 & 54.8 & 44.4 & 54 & 57.7 \\
    &\textbf{Rain-G} & 27.7 & 41.8 & 39.9 & 43.6 & 46.2 & 33.2 & 24.4 & 43.2 & 43.9 & 28.8 & 44.5 & 47.4 \\
    &\textbf{Rain-D-G} & 21.6 & 33.7 & 33.1 & 35.2 & 37.7 & 26.1 & 18.5 & 34 & 34.2 & 23 & 36.3 & 38.3 \\
    &\textbf{Defocus-G} & 32.6 & 43.3 & 40.8 & 43.4 & 45.9 & 36 & 30.0 & 43.8 & 43.4 & 32.7 & 44.9 & 47.3 \\
    \hline
    \multicolumn{2}{c|}{\textbf{Average} ($\mathrm{mS_{cor}}$)} & 40.5 & 48.7 & 48.4 & 53.0 & 54.5 & 43.4 & 41.4 & 52.5 & 52.7 & 41.9 & 51.5 & 56.0  \\
    \hline
    \end{tabular}%
}
\caption{The benchmarking results of 12 trackers on \textbf{UAV-C}. We show the performance under each corruption and the overall corruption robustness $\mathrm{mS_{cor}}$ averaged over all corruption types. The results are evaluated with Success performance defined in OPE~\cite{Wu2013OnlineOT}.}
\label{tab:results-kitti}
\end{table*}

\section{ Experiments}
\subsection{Benchmark Dataset}
UAV-C is built upon UAV-10fps and DBT70 by designing 18 corruptions, each with three varying degrees of severity. Notably, for each sequence, these corruptions are applied continuously from the first frame to the last, with the Zoom Blur effect thoughtfully implemented in a seamless zoom-in and zoom-out manner. This rigorous process has resulted in the creation of a total of 51 corrupted versions for each sequence encapsulated within the UAV-C dataset. Eventually, UAV-C contains over 9843 videos for assessment of UAV tracking under corruptions.

\subsection{Evaluated Trackers}
To understand the performance of existing methods on UAV-C and to provide baseline for future comparison, we assess 12 representative generic and UAV tackers, which can be categorized into three types: \textbf{CNN-based trackers} that implement tracking using only CNN architecture, including SiamAPN++~\cite{cao2021siamapn++}, SiamBAN~\cite{chen2020siamese}, SiamRPN++~\cite{li2019siamrpn++}, and KeepTrack~\cite{mayer2021learning}, KeepTrack is not Transformer-based, also need to update the table. \textbf{CNN-Transformer-based} that adopt hybrid CNN and Transformer architectures to achieve tracking, including ToMP~\cite{mayer2022transforming}, HIFT~\cite{cao2021hift}, TCTrack~\cite{cao2022tctrack}, TransT~\cite{chen2021transformer}, STARK~\cite{Yan2021LearningST}, and SGD~\cite{Yao2023SGDViTSD}. \textbf{Transformer-based} that employ a pure Transformer network for tracking, including Aba-ViTrack~\cite{Li_2023_ICCV} and OSTrack~\cite{ye2022joint}. Note, the reason to include generic trackers is because they are also evaluated on UAV videos. Besides, all trackers are evaluated as they are without modifications.

\begin{figure*}
    \centering
        \includegraphics[width=\linewidth]{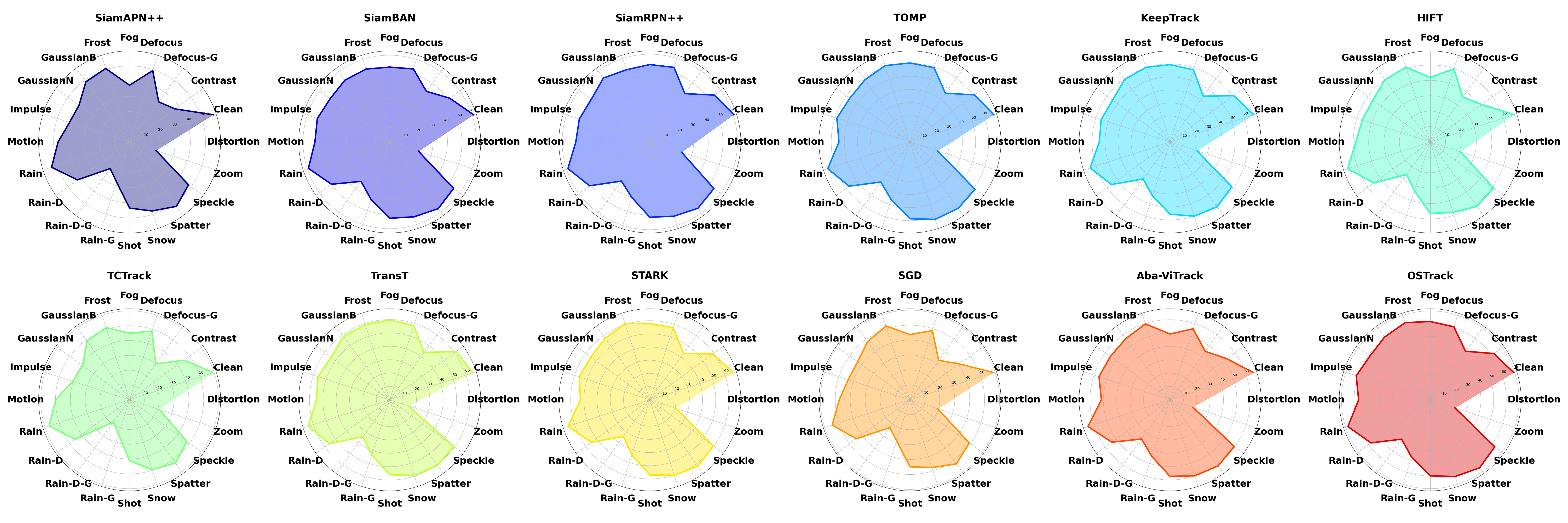}
    \caption{Mean success rate of different corruptions for each tracker. Please zoom in and view in color.}
    \label{fig:eachtracker-mean-successrate}
\end{figure*}

\subsection{Evaluation Metric}
The evaluation metric is Mean Success ($mS$) with a Success Rate defined in One Pass Evaluation (OPE) \cite{Wu2013OnlineOT}. The success measures the ratios of successful frames at the thresholds varied from 0 to 1. We design $mS_{cor}$ and $mS_{c}$ to measure the mean Success rate of all severity levels for all corruption or specific corruption, respectively. Formally, $mS_{cor}$ and $mS_{c}$ are formulated as:
\begin{small}
\begin{align}
mS_{cor} &= \frac{1}{|C|}\frac{1}{|I(c)|} \sum_{c \in C} \sum_{i \in I(c)} S_{c,i} \\
\quad mS_{c} &= \frac{1}{|I(c)|}  \sum_{i \in I(c)} S_{c,i}
\end{align}
\end{small}
where $i$ denotes the severity level and $c$ the corruption type. $S_{c, i}$ is the success rate for each corruption type c at each severity $i$.

To assess robustness of different models under all corruptions or a specific corruption, we use $rDrop_{\text{success}}^{cor}$, $rDrop_{\text{success}}^{c}$, $rDrop_{\text{precision}}^{cor}$, and $rDrop_{\text{precision}}^{c}$ to measure the relative degradation in success and precision compared to clean data, using our benchmark as a reference. Specifically, these metrics are defined as follows,
\begin{small}
\begin{align}
rDrop_{\text{success}}^{cor} &= 1 - \frac{mS_{cor}}{S_{\text{clean}}}, \quad rDrop_{\text{success}}^{c} = 1 - \frac{mS_{c}}{S_{\text{clean}}} \\
rDrop_{\text{precision}}^{cor} &= 1 - \frac{mS_{cor}}{S_{\text{clean}}}, \quad rDrop_{\text{precision}}^{c} = 1 - \frac{mS_{c}}{S_{\text{clean}}}
\end{align}
\end{small}
Here, $S_{\text{clean}}$ denotes the model's success performance on the original dataset. The terms $rDrop_{\text{success}}^{cor}(\%)$ and $rDrop_{\text{precision}}^{cor}(\%)$ quantify the overall impact of all corruptions, while $rDrop_{\text{success}}^{c}(\%)$ and $rDrop_{\text{precision}}^{c}(\%)$ measure the impact of specific types of corruption.

\renewcommand{\arraystretch}{1.15}
\begin{table*}[!t]\small
\centering
\resizebox{\linewidth}{!}{
\begin{tabular}{@{}c|ccccccccc@{}}
\hline
Distortion & Low Contrast & Defocus-G & Defocus & Fog & Frost & Gaussian Blur & Gaussian & Impulse & Motion \\
\hline
$rDrop_{\text{success}}^{c}(\%)$ & 17.7 & 36.3 & 13.8 & 18.4 & 10.9 & 13.7 & 20.3 & 18.5 & 18.2 \\
$rDrop_{\text{precision}}^{c}(\%)$ & 15.1 & 32.0 & 9.4 & 16.1 & 9.9 & 9.9 & 17.2 & 15.2 & 9.2 \\
\hline
% Second part of the table
\hline
Distortion & Rain & Rain-D & Rain-D-G & Rain-G & Shot Noise & Snow & Spatter & Speckle & Zoom \\
\hline
$rDrop_{\text{success}}^{c}(\%)$ & 3.3 & 19.1 & 51.3 & 39.1 & 17.8 & 12.2 & 7.9 & 12.2 & 68.0 \\
$rDrop_{\text{precision}}^{c}(\%)$ & 2.5 & 14.5 & 47.9 & 35.9 & 14.4 & 9.9 & 6.1 & 9.3 & 72.9 \\
\hline
\end{tabular}}
\caption{Each distortion across all trackers. We can see that the zoom corruption has the most severe impact on the tracking performance. In addition, the composite corruptions in general heavily degrade the tracking performance.}
\label{tab:tracker-relative-drop}
\end{table*}

\renewcommand{\arraystretch}{1.15}
\begin{table*}[!t]\scriptsize
\centering
\centering
\resizebox{0.85\linewidth}{!}{
\begin{tabular}{@{}c|cccccc@{}}
\hline
Tracker & SiamAPN++ & SiamBAN & SiamRPN++ & ToMP & KeepTrack  & HIFT \\
\hline
$rDrop_{\text{success}}^{cor}(\%)$ & 28.9 & 19.4 & 19.1 & 18.1 & 21.0 & 23.4  \\
$rDrop_{\text{precision}}^{cor}(\%)$ & 23.9 & 15.8 & 19.1 & 15.9 & 19.0 & 20.6  \\
\hline
% Second part of the table
\hline
Tracker& TCTrack & TransT & STARK & SGDViT & Aba-ViTrack & OSTrack \\
\hline
$rDrop_{\text{success}}^{cor}(\%)$ & 29.2 & 19.2 & 20.3 & 27.9 & 20.9 & 18.4 \\
$rDrop_{\text{precision}}^{cor}(\%)$ & 25.0 & 16.5 & 19.0 & 24.8 & 19.0 & 17.2 \\
\hline
\end{tabular}}
\caption{Robustness drop across all trackers. We can observe that the trackers are degraded in both success and precision.}
\label{tab:tracker-robustness-drop-transposed}
\end{table*}

\subsection{Main Result}

\noindent
\textbf{Overall results.} Tab.~\ref{tab:tracker-robustness-drop-transposed} shows the performance under each corruption and the overall corruption robustness $\mathrm{mS_{cor}}$ averaged over all corruption types. From Tab.~\ref{tab:tracker-robustness-drop-transposed}, we can see that all trackers are significantly degraded in presence of corruptions. For example, the performance of state-of-the-art OSTrack is dropped from 69.3\% to 56.0\% in $\mathrm{mS_{cor}}$. In the future, more efforts are needed to improve tracking performance. In addition, we observe that, while some trackers are generally more affected by certain types of corruptions than others, each tracker has its unique performance profile across the spectrum of corruptions tested, as illustrated in the normalized radar graph Fig.~\ref{fig:eachtracker-mean-successrate}. 

% \begin{figure*}[!t]
%     \centering
%     \begin{subfigure}[b]{0.33\linewidth}
%         \includegraphics[width=\linewidth]{radar_charts/radar_success_siamese.png}
%         \caption{}
%         \label{fig:sub1}
%     \end{subfigure}
%     \hfill
%     \begin{subfigure}[b]{0.33\linewidth}
%         \includegraphics[width=\linewidth]{radar_charts/radar_success_CNN_Transformer.png}
%         \caption{}
%         \label{fig:sub2}
%     \end{subfigure}
%     \hfill
%     \begin{subfigure}[b]{0.33\linewidth}
%         \includegraphics[width=\linewidth]{radar_charts/radar_success_OneStream_Transformer.png}
%         \caption{}
%         \label{fig:sub3}
%     \end{subfigure}
%     \caption{Relative success drop of Siamese tracker (a), CNN-Transformer-based trackers (b), and One-Stream Transformer trackers (c).}
%     \label{fig:whole}
% \end{figure*}

\vspace{0.3em}
\noindent
\textbf{Comparison of corruption types.}  In evaluating the different types of corruptions, ``Zoom Blur'' emerges as the most challenging, with a notable average precision drop of 72.9\% across all trackers, as detailed in Table \ref{tab:tracker-relative-drop}. This significant reduction highlights the difficulties trackers face with sudden and large-scale changes in the visual field, undermining their ability to maintain accurate tracking.

Composite corruptions, which involve multiple distortion types simultaneously, also pose a considerable challenge, leading to an overall success rate reduction between 19.1\% and 51.3\%. This wide range underscores the complexity and variability of these scenarios, suggesting that current tracking algorithms may not be fully equipped to manage such compounded distortive effects effectively.

Conversely, corruptions like ``Rain'' and ``Spatter'' have a relatively minor impact on tracking performance, with average precision drops of 2.5\% and 6.1\% respectively, as in Tab.~\ref{tab:tracker-relative-drop}. These findings indicate that most trackers can well handle these types of disturbances, maintaining a higher level of precision despite the presence of such corruptions.

\vspace{0.3em}
\noindent
\textbf{Comparison of different trackers.} We show comparison of different trackers against corruptions, as in Tab.~\ref{tab:tracker-robustness-drop-transposed}, using success drop. From Tab.~\ref{tab:tracker-robustness-drop-transposed}, we can see that OSTrack shows the relatively smallest performance drop compared to others, which, together with the overall performance in Tab.~\ref{tab:results-kitti}, shows excellent ability of OSTrack in dealing with corruptions. It is worth noting that, OSTrack is developed using pure Transformer architecture, which indicates the more powerful features can effectively help with resisting corruptions.

\subsection{Discussion}
As shown in our study, dealing with corruptions is crucial for the robustness of UAV object tracking. In order to mitigate the effects of corruption and improve model reliability, one promising approach is the utilization of corruption contrastive learning. By enhancing the similarity between clean samples and their corruption-perturbed counterparts, this method aims to strengthen the model's resilience against common distortions. This might include considering the application of contrastive learning to enhance feature representations or the development of new loss functions that better capture the corruption robustness of tracking models.

Furthermore, addressing temporal accumulation of mistakes in videos, as observed in the failure case (see Fig.~\ref{fig:teaser}) of misidentifying the target objects, could involve examining temporal consistency and continuity in feature representations. Technologies that ensure smoother transitions and more stable tracking over time such as temporal transformer could be helpful.

\section{Conclusion}

In this work, we propose a novel benchmark UAV-C that is specially designed for investigating the robustness of UAV tracking systems against common corruptions. In order to understand existing trackers on UAV-C, we conduct extensive experiments by assessing 12 representative trackers. Results show that current UAV trackers will be degraded in presence of corruptions, and more efforts are desired to improve UAV tracking robustness. Some of our insights include: (a) The \emph{zoom} and composite corruptions have more severe impact on the UAV tracking performance compared with other types; (b) The usage of Transformer can help with resistance again the corruptions for trackers and it is a promising direction to explore more powerful Transformer architecture for UAV tracking; (c) There is a discernible trend where trackers with superior performance under standard conditions tend to maintain higher robustness when confronted with adverse scenarios, showing consistency. Through UAV-C, along with our extensive analysis, we hope to draw more attention to the study of tracking robustness under corruptions.

%%%%%%%%% REFERENCES
{\small
\bibliographystyle{ieee_fullname}
\bibliography{egbib}
}

\end{document}